\documentclass{ifacconf}
\usepackage{graphicx}      
\usepackage{natbib}        
\usepackage{amsmath,amssymb,amsfonts, comment,optidef}
\usepackage{algorithmic}
\usepackage{graphicx}
\usepackage{textcomp}
\usepackage{xcolor}
\usepackage{soul} 
\usepackage{relsize}  
\usepackage{epstopdf} 
\begin{document}
\begin{frontmatter}

\title{Physics constrained learning of stochastic characteristics \thanksref{footnoteinfo}} 

\thanks[footnoteinfo]{This work was supported by Clemson University's Virtual Prototyping of Autonomy Enabled Ground Systems (VIPR-GS), a US Army Center of Excellence for modeling and simulation of ground vehicles, under Cooperative Agreement W56HZV-21-2-0001 with the US Army DEVCOM Ground Vehicle Systems Center (GVSC). \textbf{DISTRIBUTION STATEMENT A. Approved for public release; distribution is unlimited. OPSEC\# 9873}}

\author[First]{Pardha Sai Krishna Ala} 
\author[First]{Ameya Salvi} 
\author[First]{Venkat Krovi}
\author[First]{Matthias Schmid}

\address[First]{Department of Automotive Engineering, Clemson University, 
	Greenville, SC 29607 USA (e-mail: pala, asalvi, vkrovi, schmidm @clemson.edu).}

\begin{abstract}        
Accurate state estimation requires careful consideration of uncertainty surrounding the process and measurement models; these characteristics are usually not well-known and need an experienced designer to select the covariance matrices. An error in the selection of covariance matrices could impact the accuracy of the estimation algorithm and may sometimes cause the filter to diverge. Identifying noise characteristics has long been a challenging problem due to uncertainty surrounding noise sources and difficulties in systematic noise modeling. Most existing approaches try to identify unknown covariance matrices through an optimization algorithm involving innovation sequences. In recent years, learning approaches have been utilized to determine the stochastic characteristics of process and measurement models.
We present a learning-based methodology with different loss functions to identify noise characteristics and test these approaches' performance for real-time vehicle state estimation.

\end{abstract}


\end{frontmatter}

\section{Introduction}
State estimation is crucial in various engineering applications, including control systems, robotics, autonomous navigation, signal processing, and finance. The Kalman filter (KF) and its variants (Extended Kalman Filter (EKF), Unscented Kalman Filter (UKF), and Particle Filter (PF)) have been widely used for state estimation in stochastic dynamic systems. The optimal estimates for linear systems (including time-varying) require accurate knowledge of measurement and process noise covariance matrices ({\large R and Q}). However, in real-world applications, covariance matrices are generally unknown or are approximated. An error in the selection of covariance matrices could cause the filter to underestimate or overestimate the process model or measurement model, leading to suboptimal performance or even divergence of the filter, necessitating the identification of stochastic features as a key component of adaptive filtering [\cite{agamennoniApproximateInferenceStateSpace2012}]. Addressing this challenge has been the subject of extensive research over the past decades, leading to the development of various adaptive filtering techniques for noise covariance estimation.
\begin{figure}[h!]
    \centering
    \includegraphics[width=0.90\linewidth]{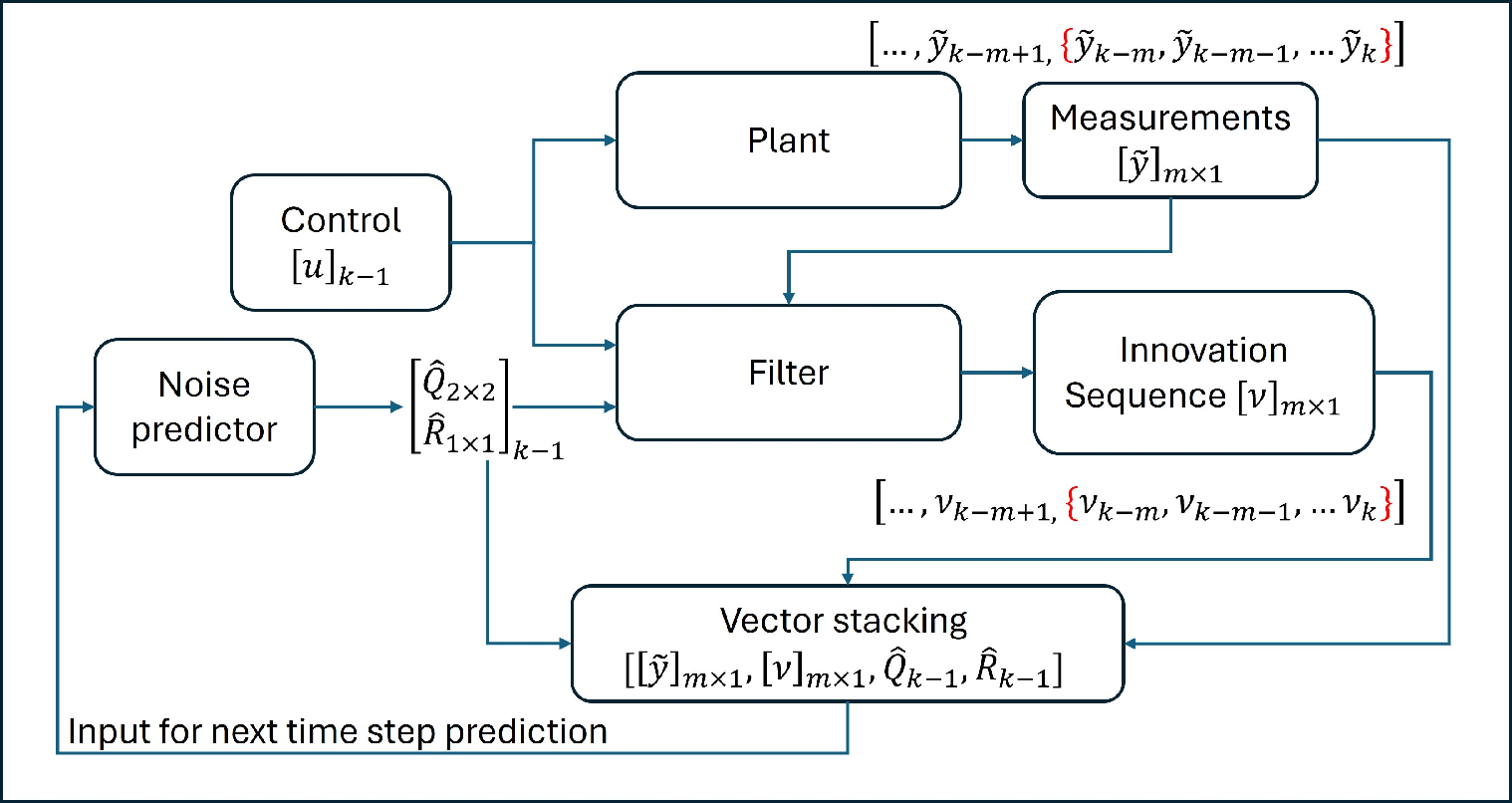}
    \caption{Implementing the noise predictor model as part of the recursive state estimation framework.}
    \label{fig:ML-NoisePredcitior} 
\end{figure}

Traditional adaptive filtering techniques such as Bayesian inference[\cite{chenKalmanFilterRobot2012}], Maximum Likelihood Estimation (MLE), correlation methods 
[\cite{mehraApproachesAdaptiveFiltering1972}, \cite{belangerEstimationNoiseCovariance1972}] use optimization-based methods to identify Q and R matrices by using the innovation sequence or measurement residuals. However, these approaches suffer from computational complexity and local optimum convergence. 

Recently, Machine learning (ML) has emerged as a promising alternative to traditional adaptive filtering for real-time identification of unknown parameters. Deep Reinforcement Learning (DRL) methods learn optimal decision-making policies through interaction with an environment, making them particularly suited for adaptive filtering problems where noise statistics change dynamically. Unlike traditional methods, which rely on analytical noise models, learning-based adaptive filtering continuously adjusts Q and R based on real-time feedback, improving estimation accuracy and robustness. To this end, a machine learning augmented solution for predicting the process and measurement noise has been proposed in this paper. Fig.~\ref{fig:ML-NoisePredcitior} illustrates the framework where a machine learning model predicts the process and the measurement noises based on the most recent m measurements.
The key contributions of this paper are:
\begin{itemize}
	\item A learning-based adaptive filtering framework for vehicle state estimation, where a framework identifies the elements of covariance matrices to minimize the estimation error.
        \item Implementing physics-constrained loss functions capturing attributes such as filter consistency to develop generalizable policies. 
\end{itemize}

The remainder of this paper is organized as follows: Section 2 reviews existing adaptive filtering methods, including traditional and ML-based approaches. Section 3 presents the proposed learning-based adaptive filtering framework, including state representation, loss function design, and training procedures. Section 4 provides simulation results and performance comparisons, while Section 5 concludes the paper and discusses future research directions.

\section{Literature Review}

In the presence of Gaussian measurement and process noises, the Kalman filter (KF) serves as the optimal state estimator for linear dynamic systems. Kalman filter is the best linear state estimator for linear systems excited by a non-Gaussian random process with known covariance [\cite{kalmanNewApproachLinear1960}]. For a linear system with known model equations and statistical properties of noise, state estimates and state error covariance can be identified. Often, in real-life applications, system dynamics and statistical properties of the noise are not entirely known, which can affect the accurate identification of the statistical properties of the state. Uniquely estimating process and measurement noise covariance matrices \textbf{Q, R} from available measurement data is necessary for adaptive filtering.

\subsection{Traditional Adaptive Filtering Approaches}
The efforts to identify methods to determine the statistical properties of measurement and process noise are not entirely new. Various approaches to estimate noise covariance matrices are classified into the following categories mentioned in  [\cite{zhangIdentificationNoiseCovariances2020}]: Maximum likelihood estimation, Bayesian estimation [\cite{husaADAPTIVEBAYESFILTERING1969}], covariance matching [\cite{1101260}], and correlation methods. Bayesian estimation 
[\cite{hilbornOptimalEstimationPresence1969}, \cite{mehraApproachesAdaptiveFiltering1972}] 
utilizes the posterior probability density function (PDF) to identify unknown elements of Q and R matrices. However, these approaches require high computational power to search through an ample parameter space, limiting their real-time applicability. 

 Early works, such as those by   
[\cite{mehraApproachesAdaptiveFiltering1972}] introduced correlation-based methods to estimate these parameters, in which the author utilizes the innovation property of the optimal filter to provide estimates for noise covariance matrices.

 Most of the proposed works can not be implemented in real-time as they involve an optimization algorithm to converge to a solution based on the measurement data. Unlike simulation, only measurement data is available in real applications, and additional information about the plant dynamics may not be available. 

 Multiple model approaches, such as Multiple-Model Adaptive Estimation (MMAE) and Interacting Multiple-Model (IMM), can identify unknown parameters in real-time, including the elements of covariance matrices. It is assumed that the model describing the accurate system dynamics exists among the bank of filters [\cite{salvi2024online}, \cite{bar-shalomEstimationApplicationsTracking2002}, \cite{crassidisOptimalEstimationDynamic2012}]. IMM filter, a suboptimal filter, allows for switching between models probabilistically and identifying unknown elements of covariance matrices. The drawback of this approach is that unknown elements of covariance matrices describing the system have to be present among the bank of filters, and the size of the bank of filters can increase, affecting the computational efficiency. 
\subsection{Learning-based Adaptive Filtering}
In recent years, learning-based approaches have gained popularity in identifying complex characteristics of noisy data [\cite{giles2001noisy}]. Learning-based approaches are used in finance and signal processing to identify noisy time series data and make predictions. Recurrent Neural Networks (RNNs) and Long Short-Term Memory (LSTMs) capture temporal dependencies in noise statistics. However, these methods require extensive training data and have limited real-time applicability. In [\cite{abbeel2005discriminative}], the authors propose a discriminative learning approach to estimate Q and R matrices to improve state estimation accuracy. In [\cite{9198927}], authors propose a deep-learning-based adaptive Kalman filter using the Mean squared error (MSE) based loss function to estimate the Q and R matrices. The drawback of this approach is that it requires offline training but offers improved computational cost.

\section{Mathematical framework}
The following section details the system dynamics and measurement model of discrete systems. 
\subsection{System Modeling}
Consider the discrete linear dynamic system:
\begin{equation}
	x_{k+1} = F_kx_k + B_ku_k + \Gamma_k w_k
        \label{Filter model}
\end{equation}
\begin{equation}
	z_k = H_kx_k +   v_k
    \label{measurement model}
\end{equation}
\begin{equation}
	w_k \sim \mathcal{N}(0, Q_k) ,
	v_k \sim \mathcal{N}(0, R_k)
        \label{eq: Covariance matrices}
\end{equation}
where \( x_k \) is an \( n \)-dimensional state vector, \( F_k \) is the state transition matrix of the system, \(u_k\) is a \(r\)-dimensional control input vector, \(B_k\) is the \(n \times r \) control input gain matrix, \( H_k \) is the \( m \times n \) measurement matrix, and \( \Gamma \) is the \( n \times w \) dimensional noise gain matrix. The sequences \( w_k) \), \( k = 0, 1, \dots \) and \( v_k \), \( k = 0, 1, \dots \), are zero-mean white Gaussian noises with covariance matrices \( Q_k \) and \( R_k \), respectively. The two noise sequences and the initial state are assumed to be mutually independent. The matrices \( Q_k \) and \( R_k \) are assumed to be positive definite. 
\subsection{Kalman Filter Modeling}
Given the estimate \( \hat{x}_k^+ \), the filter  equations describing the state estimates and their error covariances are based on [\cite{kalmanNewApproachLinear1960}]
\begin{equation}
	\tilde{x}_{k+1}^- = x_{k+1} - \hat{x}_{k+1}^-	\label{Predicted state error}
\end{equation}
\begin{equation}
	\mathlarger{\mathlarger{\nu}}_{k+1} =  H_k\tilde{x}_{k+1}^- + v_{k+1} 
    \label{innovation}
\end{equation}
\begin{equation}
	S_{k+1} = H_{k+1} P_{k+1}^- H_{k+1}^T + R_{k+1} \label{innovation covariance}
\end{equation}
\begin{equation}
	W_{k+1} = P_{k+1}^- H_{k+1}^T S_{k+1}^{-1} 
    \label{Kalman gain}
\end{equation}
where
\( \tilde{x}_{k+1}^- \) is the one-step predicted state error, 
\( W_{k} \), \( k = 1, \dots, N \) is the sequence of Kalman filter gains,
\( \mathlarger{\mathlarger{\nu}}_{k} \), \( k = 1, \dots, N \) is the innovation sequence,
\( S_{k + 1} \) is the measurement prediction (or innovation) covariance,

When the dynamical system is linear time-invariant, and steady-state Kalman gain is determined, state error at $\textit{k}^{\text{th}}$ instant can be expressed by recursively going \textit{m} steps backward as follows
\begin{equation}
	\begin{aligned}
		\tilde{x}_k^{-} &= \left[ F(I_{n} - W H) \right]^k \tilde{x}_{k-m}^- 
		- \sum_{j=1}^{m} \left[ F (I - W H) \right]^{j-1} F W v_{k-j} \\
		&\quad + \sum_{j=1}^{m} \left[ F (I - W H) \right]^{j-1} \Gamma w_{k-j}.
	\end{aligned}
	\label{eq: Recursive error}
\end{equation}

Using the eq. \ref{eq: Recursive error}, the innovation sequence can be expressed in state error, process noise, and measurement noise terms. The autocorrelation matrix for the innovation sequence can be expressed based on the steady-state Kalman gain, $W$, as follows
\begin{equation}
	\tilde{F} = \left[ F (I - WH), \right]
\end{equation}

\begin{equation}
	C_0 = E[\nu_k \nu_k^T] = HP_k^-H^T + R_k
        \label{Innovation covariance matrix}
\end{equation}

\begin{equation}
    C_m = E[\nu_k \nu_{k-m}^T]
    \label{cross-correlation matrices}
\end{equation}
When the steady-state error covariance $(P)$ and stationary noise \textit{(constant covariance matrix)} are used,
\begin{equation}
	\begin{aligned}
		C_m &= H \tilde{F}^{m-1} F \left[  P H^T - W C_0  \right], \quad i > 0.
	\end{aligned}
        \label{cross-correaltion_final}
\end{equation}

When optimal Kalman gain is used, $C_i$ vanishes for $ i > 0$; based on the innovation sequence, estimates for M autocorrelation matrices are defined as

\begin{equation}
	\hat{C}_i = \frac{1}{N - M} \sum_{j=1}^{N-M} \nu_j \nu_{j+i}^T, 
	\quad i = 0, 1, 2, \dots, M-1
        \label{eq:sample-correlation-matrices}
\end{equation}

\subsection{Conditions for Optimality}

For a Kalman filter to be optimal, it must satisfy bias, consistency, and efficiency conditions [\cite{crassidisOptimalEstimationDynamic2012}]. If the information about unknown plant dynamics is only known through measurement data, we could utilize statistical hypothesis testing for time-averaged autocorrelation (\ref{time-averaged autocorrelation}) and time-averaged normalized innovation square (\ref{eq:time-averaged-NIS}) statistics of innovation sequence [\cite{bar-shalomEstimationApplicationsTracking2002}].

\begin{equation}
    \left [\bar{\rho}_l(j) \right] =
    \sum_{k=1}^{N} \nu_l(k) \nu_l(k+j) 
    \left[ 
    \sum_{k=1}^{N} \nu_l(k)^2 
    \sum_{k=1}^{N} \nu_l(k+j)^2 
    \right]^{-1/2} \\
    \label{time-averaged autocorrelation}
\end{equation}

\begin{equation}
    \bar{\epsilon_{\nu}} = \frac{1}{N} \sum_{k=1}^{N} \nu(k)' S(k)^{-1} \nu(k)
    \label{eq:time-averaged-NIS}
\end{equation}

\subsection{Vehicle Model}~\label{sec:bicycle-model}
The vehicle motion can be described by different models depending on the level of fidelity required. 
In the actual plant model, which can be unknown, modeling errors can arise due to uncertainty in cornering stiffness (considering the effects of lateral load transfer and the nonlinear relationship between cornering stiffness and normal load), mass, moment of inertia, and other vehicle parameters. In our approach, the plant model is based on a bicycle model with varying cornering stiffness for the inner and outer tires of the axle. Due to the effect of lateral load transfer, cornering stiffness varies non-linearly with normal load, affecting the overall cornering stiffness of the axle. The filter model for vehicle state estimation is based on the simplified linear bicycle model (\textit{single track model}) with nominal values for cornering stiffness [\cite{rajamanivd}]. It is a 2-DoF motion model describing lateral dynamics through the states, vehicle slip angle and yaw rate [${\beta}$, $\dot{\psi}$]. A linear tire model, [$F=C*{\alpha}$], with the cornering stiffness per axle, is used to model terrain-tire interaction and describe the dynamics of vehicle motion (\ref{eq:contdyn}). The discretized version of the bicycle model with nominal cornering stiffness is used for propagation and update of estimated states 
based on the measurement data.

\begin{figure}[h!]
    \centering
    \includegraphics[width=0.90\linewidth]{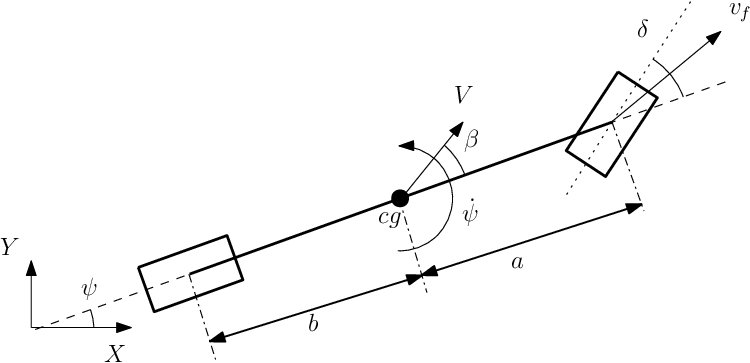}
    \caption{Bicycle model of a vehicle.}
    \label{fig:bicycmod} 
\end{figure}

\begin{align} \label{eq:contdyn}
    \underbrace{\begin{bmatrix} \nonumber
        \dot{\beta}\\
        \ddot{\psi}
    \end{bmatrix}(t)}_{\dot{x}(t)} = & \underbrace{\begin{bmatrix}
        -\frac{C_{\alpha_f}+C_{\alpha_r}}{mV} & -\frac{aC_{\alpha_f}-bC_{\alpha_r}}{mV^2} \\
        -\frac{aC_{\alpha_f}-bC_{\alpha_r}}{I_z} & -\frac{a^2C_{\alpha_f}-b^2C_{\alpha_r}}{I_zV}
    \end{bmatrix}}_F \underbrace{\begin{bmatrix}
        {\beta}\\
        \dot{\psi}
    \end{bmatrix}(t)}_{x(t)}\\ 
    & + \underbrace{\begin{bmatrix}
        \frac{C_{\alpha_f}}{m}\\
        \frac{aC_{\alpha_f}}{I_z}
    \end{bmatrix}}_B\underbrace{\delta(t)}_{u(t)} + \underbrace{\begin{bmatrix}
        1 & 0\\
        0 & 1
    \end{bmatrix}}_{\Gamma} \underbrace{\begin{bmatrix}
        w_1 \\
        w_2
    \end{bmatrix}}_w
\end{align}
\begin{equation}
	w_1 \sim \mathcal{N}(0, Q_a) ,
	w_2 \sim \mathcal{N}(0, Q_b)
        \label{eq: Covariance-matrices}
\end{equation}

\subsection{Deep Learning}

While deep learning can be utilized in a broader context, this work proposes using Long Short-Term Memory (LSTM) models as universal function approximators to predict the process and measurement noise. Once trained, the LSTM model is implemented in the real-time filtering framework (as illustrated in Fig.~\ref{fig:ML-NoisePredcitior}) to predict the noise values based on the latest $m$ measurements captured by a moving window of size $ m \times 1$.

\subsubsection{Training Dataset:} 

\begin{figure}[t]
    \centering
    \includegraphics[width=1\linewidth]{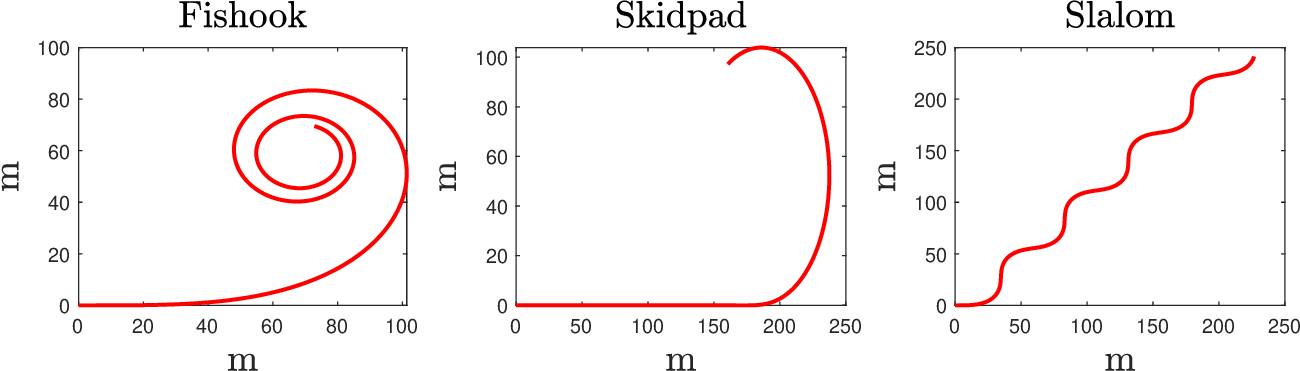}
    \caption{Three motion maneuvers (Fishook, Skidpad and Slalom) of a vehicle's CG used for collecting a dataset of 50000 training samples}
    \label{fig:ManeuverDataset}
\end{figure}

A vehicle motion dataset is collected by simulating the bicycle model outlined in section~\ref{sec:bicycle-model} by introducing random values for the process noise $Q_{2 \times 2}$ and the measurement noise $R_{1 \times 1}$. Here, the motion dataset comprises data from different maneuvers as shown in fig~\ref{fig:ManeuverDataset}, introducing levels of complexity to vehicle modeling. The skidpad (constant steering input) test, performed at constant velocity, introduces constant lateral load transfer compared to the fishhook maneuver (varying lateral load transfer). Data collection from different possible maneuvers could generate rich datasets with varying Q for training. The randomly generated values serve as true labels for the training dataset, which the model is trained to predict based on the model's input. At any timestep, $k$, the model predicts $Q_k$ and $R_k$, which are used to get the next timestep's residual measurement, $\nu_{k+1}$. Similarly, at each $(k+1)^{th}$ timestep, a measurement $\tilde{y}_{k+1}$, is received from the plant model. Both the residual measurement and the true measurements are maintained within independent memory buffers $[\nu]$ and $[\tilde{y}]$, from which a sequence of $m$ samples is collected and concatenated with the previous timestep's $Q$ and $R$, to serve as the model input for the next timestep's prediction. 

\begin{equation}
    \begin{bmatrix}
        \tilde{y}_{m \times 1} \quad \nu_{m \times 1} \quad \hat{Q}_{k-1} \quad \hat{R}_{k-1}
    \end{bmatrix}_{k}
    \rightarrow
    \begin{bmatrix}
        \hat{Q} \\
        \hat{R}
    \end{bmatrix}_{k}
\end{equation}

In this work, the length of the measurement sequence, $m$, is selected to be 100, and the noise variances have been bounded to 1e-3. Further, for training simplicity, Q is simplified as a diagonal matrix of $Q_a$ and $Q_b$ as shown in eq.~\ref{eq: Covariance-matrices}. Thus, three prediction labels ($Q_a, Q_b$ and $R$) correspond to the noise characteristics of the process noise for the two model states and one measurement.

\subsubsection{Training hyperparameters:}
Training hyperparameters can be crucial in setting up the training's sensitivity analysis baselines and providing insight for recreating the results. The model utilized in this work was trained on an NVIDIA A2000 GPU using the set of hyperparameters outlined in the table~\ref{tab:hyperparameters}.

\begin{table}
\label{tab:hyperparameters}
    \centering
    \begin{tabular}{cc}
       Parameter  & value\\
       \hline
       layers  & 2: LSTM - FC\\
       activation  & relu\\
       learning rate  & 1e-4 with ADAM\\
       batch size  & 128\\
       epochs & 25
    \end{tabular}
    \caption{Model training hyperparameters}
    \label{tab:hyperparameters}
\end{table}

\subsubsection{Loss Function:}~\label{sec:loss-function}

Loss functions define the error between the predicted and actual labels of the training dataset and are used as the guiding mechanisms to modify the model learning. The vanilla approach toward deep learning entails minimizing the norm between the actual and predicted labels. This function is illustrated in the equation~\ref{eq:LossA} where $Y$ are the actual labels and $\hat{Y}$ are the model predictions.

\begin{align}
    \mathcal{L}_{1} &= \mathcal{W}_{1} \lvert(Y - \hat{Y})\rvert ~\label{eq:LossA}\\
    \mathcal{L}_{2} &= \mathcal{W}_{1} \lvert(Y - \hat{Y})\rvert + \mathcal{W}_{2} \lvert \hat{C}_i \rvert ~\label{eq:LossB} \\
    \mathcal{L}_{3} &= \mathcal{W}_{1} \lvert(Y - \hat{Y})\rvert + \mathcal{W}_{3} \lvert \bar{\epsilon_{\nu}} \rvert ~\label{eq:LossC} \\
    \mathcal{L}_{4} &= \mathcal{W}_{1} \lvert(Y - \hat{Y})\rvert + \mathcal{W}_{2} \lvert \hat{C}_i \rvert  + \mathcal{W}_{3} \lvert \bar{\epsilon_{\nu}} \rvert ~\label{eq:LossD}
\end{align}

In recent years, modifying the loss functions with physics constraints has shown plausible performance improvement in improving the generalizability of the trained models on the unseen datasets (data not used during model training)[~\cite{jia2020pgml}]. To this end, the vanilla loss function in equation~\ref{eq:LossA} is augmented with the physics constraints introduced in the eq.~\ref{eq:sample-correlation-matrices} and eq.~\ref{eq:time-averaged-NIS}, independently and in combination. The resultant loss functions are illustrated in equations~\ref{eq:LossB} to ~\ref{eq:LossD}. For all the parameters of the loss function, $\mathcal{W}_{i}, \forall i\in (1,3)$ are the weights on the particular constraints. For all experiments, the following weights have been kept fixed to maintain consistency: $\mathcal{W}_1 = 1.0$, $\mathcal{W}_2 = 0.1$, $\mathcal{W}_3 = 0.1$.

\section{Implementation and Results}

\begin{figure}[tp]
    \centering
    \includegraphics[width=1\linewidth]{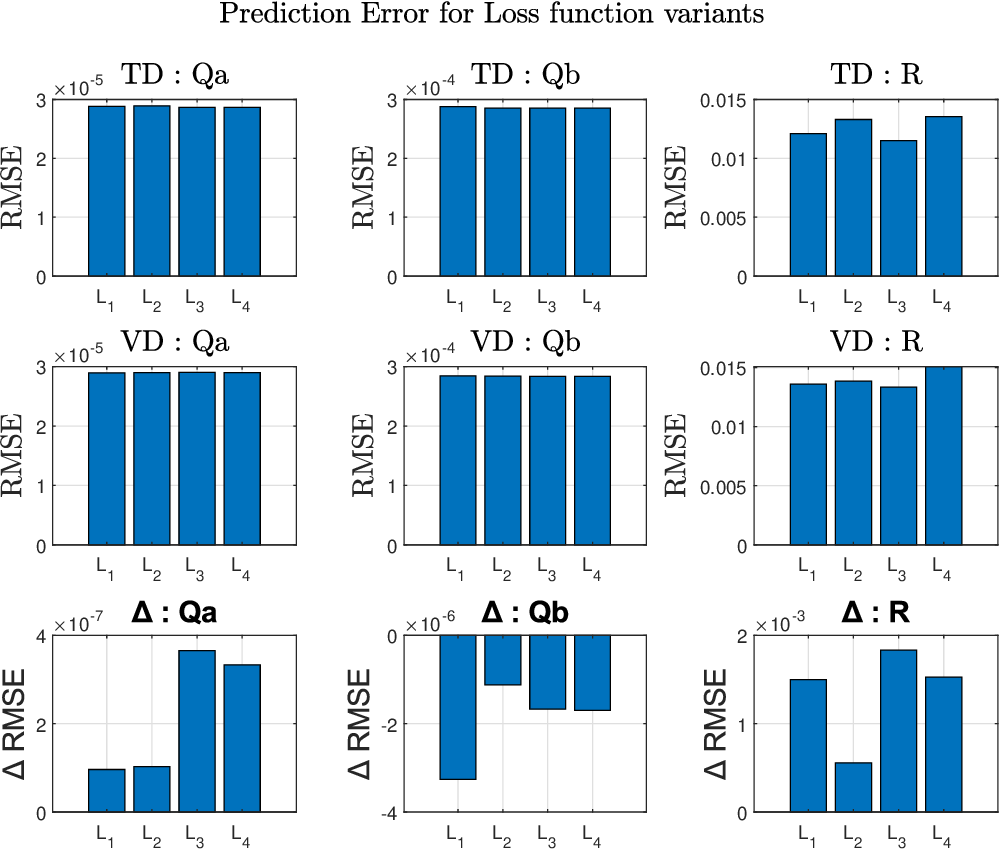}
    \caption{Label prediction errors on training and validation data for the labels $Q_a$, $Q_b$ and R on training and validation datasets (TD and VD, respectively)}
    \label{fig:BarPlots}
\end{figure}

\begin{figure*}[h]
    \centering
    \includegraphics[width=1\linewidth]{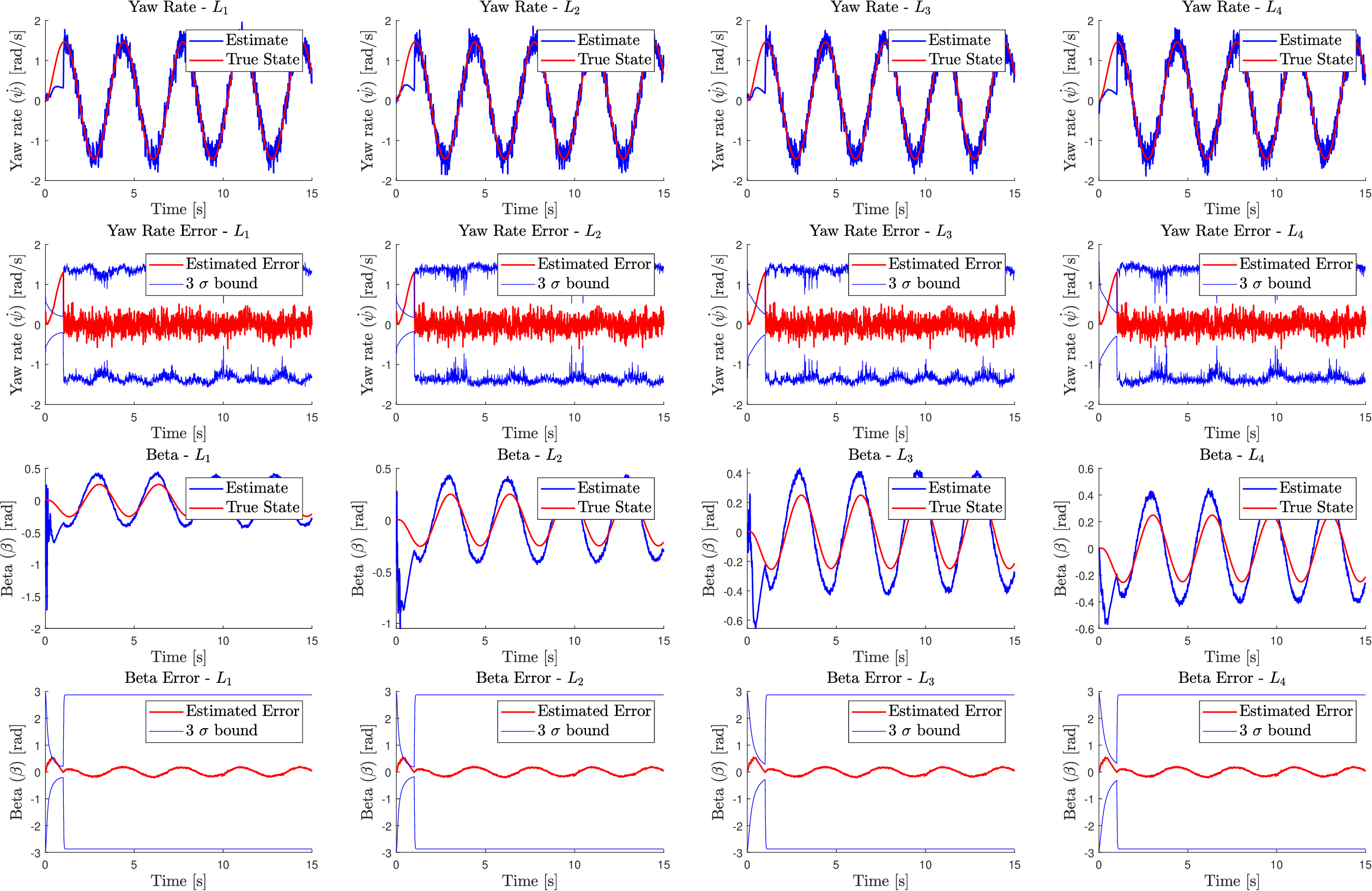}
    \caption{True states and Estimates for Yaw Rate ($\dot{\psi}$)  and vehicle slip angle ($\beta$) with error (difference between the true state and estimates) within 3$\sigma$ bounds for the Slalom vehicle maneuver.}
    \label{fig: YawRateEstimation}
\end{figure*}


\begin{figure}
    \centering
    \includegraphics[width=1\linewidth]{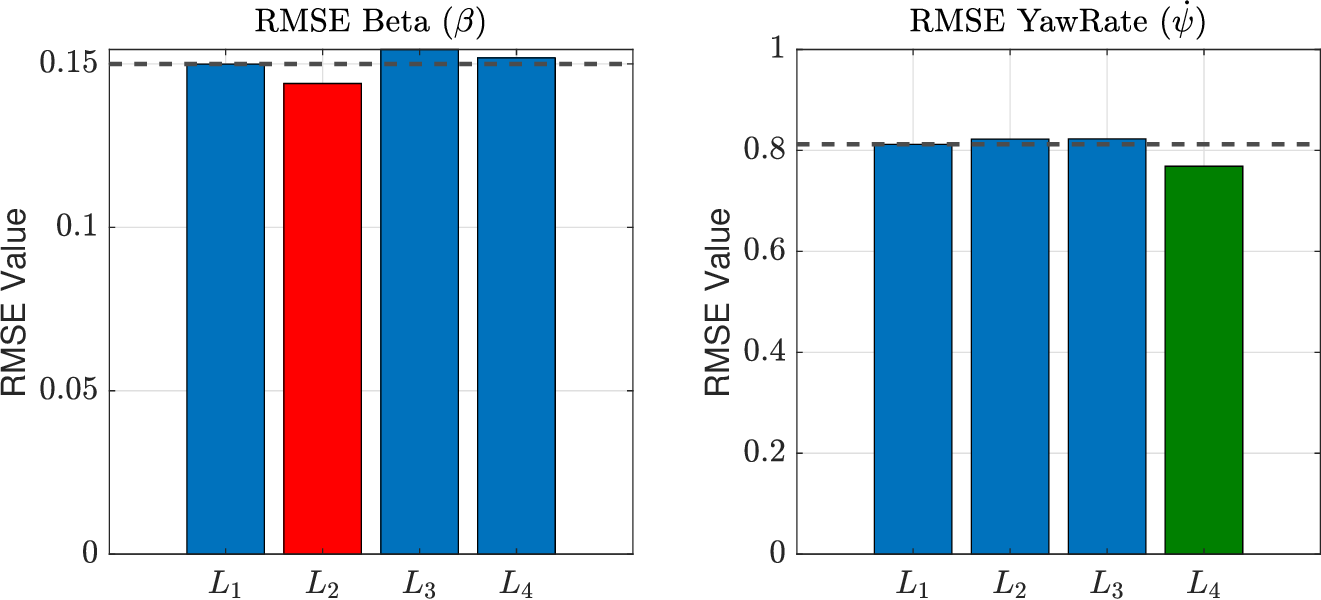}
    \caption{RMSE values for the state estimates (yaw rate and beta) for 25 random runs on validation data trajectories for variants in the loss function. Black dotted lines indicate a baseline for the $L_{1}$ loss to provide a reference for the performance of other loss function variants.}
    \label{fig:BarPlots25Random}
\end{figure}


\subsection{Training Results: Prediction Accuracy}

Four variants of the prediction model based on the four loss functions discussed in section~\ref{sec:loss-function} were trained on the training dataset. Figure~\ref{fig:BarPlots} illustrates the label prediction performance of the trained models. It can be seen that from the label prediction point of view, $\mathcal{L_2}$ performs better than all other models on all prediction labels. The last row in fig.~\ref{fig:BarPlots} shows the deviation in the prediction error between the training and validation data. While the RMSE values increased for the predictions of $Q_a$ and $R$ values, the $Q_b$ values, on the other hand, went down. In general, the RMSE values on the unseen validation data should typically go up, and the anomaly in $Q_b$ values could be due to bias in the validation dataset.

The initial hypothesis was that introducing physics constraints in the loss function should help with generalizability on the unseen dataset. While the autocorrelation-based loss function improved the prediction accuracy of the validation data for $Q_b$ and $R$ values, no other significant performance change was seen for different values. Tuning the learning hyperparameters (in particular, increasing the dataset size, changing the learning rate and adding more network layers) can be a subject of future study to investigate the impact of the physics-based loss functions.

\subsection{Training Results: State Estimation}

Fig.~\ref{fig: YawRateEstimation} shows the estimated state and state error and its 3-$\sigma$ bounds for a slalom maneuver with different loss functions. From the first row of the figure, we would observe the yaw rate estimate ($\hat{\dot{\psi}}$) to track the actual state value, but it shows the values to be fluctuating as the predicted process noise covariance ($ \hat{Q_b}$) does not entirely converge to the actual value. The convergence to actual values can be potentially improved with extended training, richer datasets and hyperparameter tuning. Across the proposed loss functions, estimates do not show a significant difference. In the 3-$\sigma$ plot, the yaw rate error lies within the 3-$\sigma$ bounds for most of the data for the predicted values of Q and R. The state error is initially out of bounds because the proposed framework requires a set number of samples to predict the initial noise covariance matrices. 


As compared to the yaw rate estimates, state estimates for vehicle slip angle ($\hat{\beta}$) show slightly subpar performance, likely due to the sole availability of the yaw rate ($\dot{\psi}$) measurements.
In particular, the vehicle slip angle estimates overshoot the true values, potentially due to poor prediction accuracy in the $Q_a$ values. A possible reason for such behavior is the inability of the learning model to capture the implicit information about $Q_a$, which is hidden within the state estimation model, compared to $Q_b$, which is available more explicitly due to the yaw rate measurements.


Fig.~\ref{fig:BarPlots25Random} shows the RMSE values of the state estimation errors for vehicle slip angle ($\beta$) and yaw rate ($\dot{\psi}$). For the $\beta$, the loss function minimizing the norm of the time-averaged autocorrelation metric ($L_2$) shows better performance than other loss functions, whereas for ($\dot{\psi}$) estimation, minimization of NIS metric and autocorrelation ($L_4$) performs better than other loss functions. The improvements in the RMSE metric indicate that the infusion of physics constraints improves the state estimation with better identification of process and measurement noise covariances. While these results point towards the potential merits of introducing physics constraints in the loss functions, the exact influence of a specific constraint (autocorrelation or NIS) is yet to be investigated.

\section{Discussion}
In this paper, we proposed physics-based loss functions to identify the characteristics of the process and measurement noise. Our preliminary investigations indicate the merits of introducing physics constraints for improving the state estimation for vehicle motion maneuvers. In the proposed approach, identifying the measurement noise covariance matrix yielded significantly better results than identifying the process noise covariance matrix, which can be attributed to the hidden nature of process noise in the measurement data. As a future study, it would be interesting to benchmark our approach against other real-time noise prediction frameworks.


For future work, we propose to investigate the impact of weights of the loss function parameters on the overall state estimation performance. Furthermore, the dataset can potentially be augmented with more maneuvers under different conditions to assess the framework's ability to predict noise covariance matrices. It would also be helpful to validate the proposed approach to identify the characteristics of process and measurement noise for time-varying and nonlinear systems. 
\bibliography{AdaptiveEstimation}             

\appendix

\end{document}